\documentclass[10pt,twocolumn,letterpaper]{article}

\usepackage[pagenumbers]{cvpr} 

\definecolor{cvprblue}{rgb}{0.21,0.49,0.74}
\usepackage[pagebackref,breaklinks,colorlinks,allcolors=cvprblue]{hyperref}

\usepackage[linesnumbered,ruled,vlined,algo2e]{algorithm2e}
\usepackage{algpseudocode}
\usepackage{makecell}
\usepackage{multirow}
\usepackage[table]{xcolor}
\usepackage{tabularx}
\usepackage{booktabs}
\usepackage{wrapfig}
\usepackage{graphicx}
\usepackage{enumitem}
\usepackage{threeparttable}
\usepackage{listings}
\usepackage{pythonhighlight}

\def\paperID{****} 
\def\confName{CVPR}
\def\confYear{2026}

\title{\textsc{CTCal}: Rethinking Text-to-Image Diffusion Models via Cross-Timestep Self-Calibration}

\author{
{\bf Xiefan Guo\textsuperscript{1,2,3}\quad 
Xinzhu Ma\textsuperscript{1,2,3}\quad 
Haiyu Zhang\textsuperscript{2,3}\quad 
Di Huang\textsuperscript{1,2}\footnotemark[1]}\\[5pt]
{\normalsize \textsuperscript{1}State Key Laboratory of Complex and Critical Software Environment, Beihang University, Beijing 100191, China}\\
{\normalsize \textsuperscript{2}School of Computer Science and Engineering, Beihang University, Beijing 100191, China}\\
{\normalsize \textsuperscript{3}Shanghai Artificial Intelligence Laboratory, Shanghai 200232, China}\\
{\tt\small \{xfguo,xinzhuma,zhyzhy,dhuang\}@buaa.edu.cn}\\
}

\begin{document}
\maketitle

\footnotetext[1]{Corresponding author.}

\begin{abstract}
Recent advancements in text-to-image synthesis have been largely propelled by diffusion-based models, yet achieving precise alignment between text prompts and generated images remains a persistent challenge. We find that this difficulty arises primarily from the limitations of conventional diffusion loss, which provides only implicit supervision for modeling fine-grained text-image correspondence. In this paper, we introduce Cross-Timestep Self-Calibration (\textsc{CTCal}), founded on the supporting observation that establishing accurate text-image alignment within diffusion models becomes progressively more difficult as the timestep increases. \textsc{CTCal} leverages the reliable text-image alignment ({i.e.}, cross-attention maps) formed at smaller timesteps with less noise to calibrate the representation learning at larger timesteps with more noise, thereby providing explicit supervision during training. We further propose a timestep-aware adaptive weighting to achieve a harmonious integration of \textsc{CTCal} and diffusion loss. \textsc{CTCal} is model-agnostic and can be seamlessly integrated into existing text-to-image diffusion models, encompassing both diffusion-based ({e.g.}, SD 2.1) and flow-based approaches ({e.g.}, SD 3). Extensive experiments on T2I-Compbench++ and GenEval benchmarks demonstrate the effectiveness and generalizability of the proposed \textsc{CTCal}. Our code is available at \url{https://github.com/xiefan-guo/ctcal}.
\end{abstract}
\section{Introduction}
\label{sec:introduction}

Text-to-image synthesis aims to generate visually realistic images that accurately reflect input text prompts. Early advances in this field were predominantly driven by Generative Adversarial Networks (GANs) \citep{zhang2017stackgan,xu2018attngan,zhu2019dm,zhang2021cross,tao2022df} and Autoregressive Models (ARs) \citep{ramesh2021zero,ding2021cogview,yu2022scaling,chang2023muse}. Recently, Diffusion Models (DMs) \citep{ho2020denoising,dhariwal2021diffusion} have emerged as the dominant paradigm, demonstrating superior capabilities in generating high-fidelity and semantically coherent images \citep{nichol2022glide,ramesh2022hierarchical,saharia2022photorealistic,rombach2022high,guo2025shortft,flux2024,esser2024scaling,zhang2025diffusion,xie2025sana}.

\begin{figure*}
\centering
\setlength{\belowcaptionskip}{-0.2cm}
\setlength{\abovecaptionskip}{0.2cm}
\includegraphics[width=0.935\linewidth]{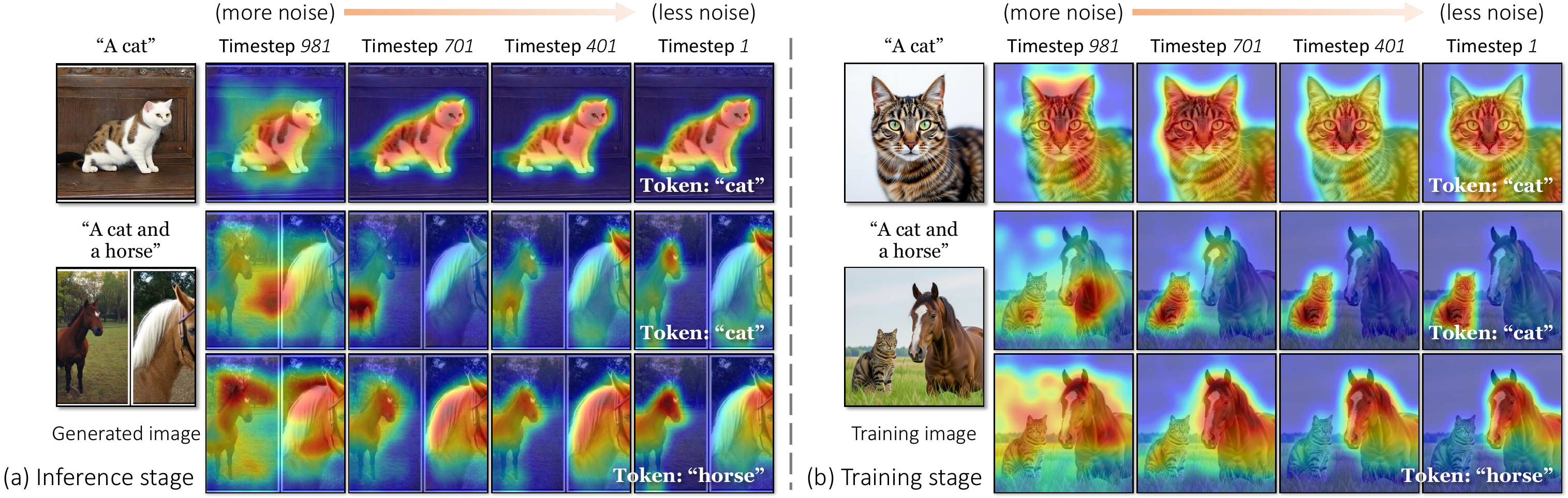}
\caption{\textbf{Investigation on the cross-attention maps.} \textbf{(a) Inference stage.} In line with existing inference-time optimization methods \citep{hertz2023prompt,chefer2023attend}, we delve into the analysis of cross-attention maps produced during the inference stage of the text-to-image diffusion model. Notably, satisfactory text-image correspondences are established for simple text prompts. Nevertheless, with more intricate text prompts, the prevalent method encounters challenges in precisely mapping the target semantics to the correct spatial position, leading to semantically inconsistent images. \textbf{(b) Training stage.} Given the text-image-noise triplet, we gather cross-attention maps at varied timesteps in training mode. A noteworthy finding emerges: cross-attention maps obtained at smaller timesteps exhibit substantially better alignment with the ground-truth image structure and semantics, while this alignment substantially deteriorates at larger timesteps. This suggests that the conventional diffusion loss, which is ubiquitously employed in current training protocols, is effective primarily at smaller timesteps. Moreover, this inability to establish precise alignments at larger timesteps, \emph{i.e.}, initial stage of inference process, constitutes a critical bottleneck, fundamentally constraining the overall fidelity and semantic accuracy of text-to-image generation.}
\label{fig:b_attn_map_vis}
\end{figure*}

To further enhance the image quality and faithfulness to text prompts, researchers have introduced numerous architectural innovations, including flow-based mechanisms \citep{albergo2023building,lipman2023flow,liuf2023low,ma2024sit}, Diffusion Transformer (DiT) \citep{peebles2023scalable,chen2024pixart,li2024hunyuan}, Multi-Modal Diffusion Transformer (MM-DiT) \citep{esser2024scaling,flux2024}, {\it etc.} Despite these strides, achieving precise and reliable alignment between text prompts and generated images remains an open challenge, especially for complex text prompts, primarily due to limitations in modeling fine-grained text-image correspondence \citep{hertz2023prompt,chefer2023attend} (see Fig.~\ref{fig:b_attn_map_vis} (a)).

Both the cross-attention layer and MM-DiT play the pivotal role in modeling the relationship between text prompts and images, contributing to text-conditioned guidance. These components are typically optimized within existing text-to-image diffusion models utilizing the conventional diffusion loss. However, this implicit approach for learning the text-image correspondence proves to be inadequate for capturing complex correspondences, particularly for larger timesteps with more noise, ultimately impairing the fidelity of the synthesized images.

Existing inference-time optimization methods \citep{li2023divide,chefer2023attend,xie2023boxdiff,kim2023dense,phung2024grounded,guo2024initno} typically explore the evolution of text-image correspondence (\emph{i.e.}, cross-attention maps) during inference, and suffer from limited generalizability and scalability. In this work, we rethink that from the perspective of the training phase: the challenge of learning the text-image correspondence within text-to-image diffusion models escalates with the progression of timesteps, transitioning from simple to complex scenarios. Empirically, as shown in Fig.~\ref{fig:b_attn_map_vis} (b), cross-attention maps extracted at smaller timesteps with less noise aligns more accurately with the provided image and corresponds more closely to the semantic distribution in the spatial dimension. This implies the denoising network handles text-image correspondence more effectively under conventional diffusion loss at smaller timesteps. However, this task becomes increasingly difficult at larger timesteps.

Drawing from these findings, we introduce Cross-Timestep Self-Calibration (\textsc{CTCal}), a fine-tuning method that capitalizes on the robust text-image alignment (\emph{i.e.}, cross-attention maps) established at smaller timesteps to calibrate the learning at larger timesteps, achieving explicit self-supervision. Moreover, we propose a part-of-speech-based cross-attention map selection strategy, prioritizing the attention maps corresponding to the noun tokens that contribute most directly to spatial comprehension and eliminating noise interference. We introduce pixel-semantic space joint optimization to augment guidance performance and propose subject response alignment regularization to counteract the potential performance degradation due to unequal subject (noun) response. We achieve a harmonious integration of \textsc{CTCal} and diffusion loss using a timestep-aware adaptive weighting.

\textsc{CTCal} is model-agnostic and can be seamlessly integrated into existing text-to-image diffusion models, including both diffusion-based (\emph{e.g.}, SD 2.1) and flow-based approaches (\emph{e.g.}, SD 3). Comprehensive evaluations on T2I-Compbench++ \citep{huang2025t2i} and GenEval \citep{ghosh2023geneval} benchmarks demonstrate the effectiveness and generalizability.

\section{Preliminaries}
\label{sec:preliminaries}

This section presents a brief review of text-to-image diffusion models, cross-attention layer, and multi-modal diffusion transformer, the latter two being instrumental in modeling text-image correspondence and actualizing text-conditioned guidance.

\begin{figure*}
\centering
\setlength{\belowcaptionskip}{-0.2cm}
\setlength{\abovecaptionskip}{0.2cm}
\includegraphics[width=0.875\linewidth]{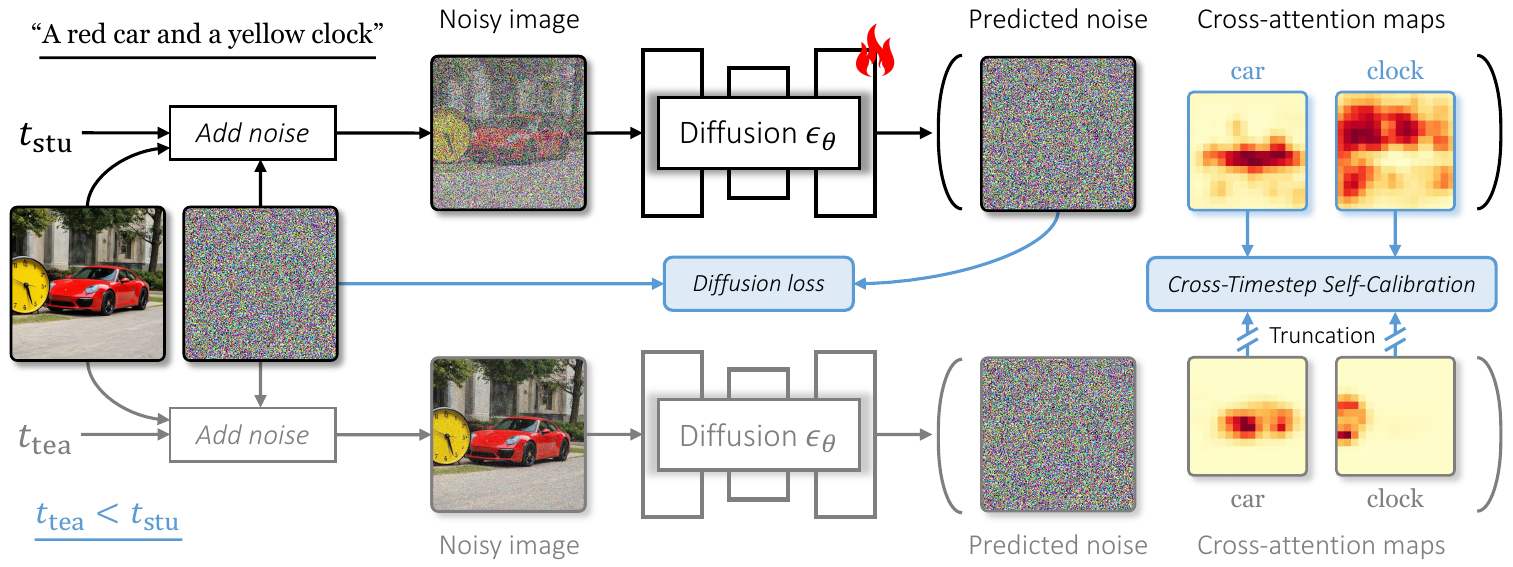}
\caption{\textbf{Illustration of \textsc{CTCal}.} \textsc{CTCal} is dedicated to leverage the reliable text-image alignment established at smaller timesteps ($t_{\text{tea}}$) to calibrate the learning process at larger timesteps ($t_{\text{stu}}$). This approach provides explicit supervision for the modeling of text-image correspondence, thereby enhancing the overall performance of text-to-image generation. Notably, the two diffusion models share identical parameters, which is solely for the convenience of presentation.}
\label{fig:ctcal}
\end{figure*}

\noindent \textbf{Text-to-image diffusion models.} Given an image $\mathbf{I}_{\text{real}}$, a text prompt $\mathbf{y}$, a Gaussian noise $\mathbf{\epsilon}$, and a timestep $t$, the text-to-image diffusion model $\epsilon_\theta(\cdot)$ is optimized with the following diffusion loss:
\begin{equation}
\label{eq:t2i-diffusion}
   \mathcal{L}_{\text{diffusion}} = \mathcal{D}\left( \mathbf{\epsilon}, \epsilon_{\theta}\left( \mathtt{Add\_Noise}\left( \mathbf{I}_{\text{real}}, \mathbf{\epsilon}, t \right), \mathbf{y}, t \right) \right),
\end{equation}
where $\mathtt{Add\_Noise}(\cdot)$ denotes the add noise function and $\mathcal{D}(\cdot)$ is a distance metric, typically implemented as a weighted mean squared error. Although existing text-to-image diffusion models such as SD 2.1 \citep{rombach2022high}, SD 3 \citep{esser2024scaling}, and FLUX.1 \citep{flux2024} differ in their specific noise addition and loss formulations, they uniformly adhere to this paradigm.

\noindent \textbf{Cross-attention layer.} The cross-attention layer is employed to establish the text-image correspondence in classic text-to-image diffusion models \citep{rombach2022high,podell2024sdxl}. Formally, the feature $f_{\text{image}}\left( \mathbf{z} \right)$ extracted from the noisy image $\mathbf{z}$ is projected to the query $\mathbf{Q} = \mathcal{Q}\left( f_{\text{image}}\left( \mathbf{z} \right) \right)$, while the text embedding $f_{\text{text}}\left(\mathbf{y}\right)$ encoded with the provided text prompt $\mathbf{y} = \{\mathbf{y}_1, \mathbf{y}_2, \cdots, \mathbf{y}_n \}$ is projected as the key $\mathbf{K} = \mathcal{K}\left( f_{\text{text}}\left(\mathbf{y}\right) \right)$ and the value $\mathbf{V} = \mathcal{V}\left( f_{\text{text}}\left(\mathbf{y}\right) \right)$, with $\mathcal{Q}\left(\cdot\right)$, $\mathcal{K}\left(\cdot\right)$, and $\mathcal{V}\left(\cdot\right)$ denoting the linear projections. The cross-attention map $\mathbf{A}$ is computed as: $\mathbf{A} = \text{softmax}\left( \frac{\mathbf{Q}\mathbf{K}^T}{\sqrt{d}} \right)$, where $d$ is channel dimension. For ease of representation, we omit the denoising timestep $t$. We denote the cross-attention map that corresponds to the $i$-th text token as $\mathbf{A}_{\mathbf{y}_i}$.

\noindent \textbf{Multi-modal diffusion transformer (MM-DiT).} Advanced text-to-image diffusion models \citep{esser2024scaling,flux2024} introduce the MM-DiT, which diverges from conventional diffusion models by concatenating text and image token embeddings into a unified input sequence. This sequence is then processed by transformer modules that utilize a joint self-attention layer. Formally, MM-DiT is formulated as: $\mathbf{Q} = \text{Concat}\left( \mathcal{Q}_{\text{image}}\left( f_{\text{image}}\left( \mathbf{z} \right) \right), \mathcal{Q}_{\text{text}}\left( f_{\text{text}}\left( \mathbf{y} \right) \right) \right)$, $\mathbf{K} = \text{Concat}\left( \mathcal{K}_{\text{image}}\left( f_{\text{image}}\left( \mathbf{z} \right) \right), \mathcal{K}_{\text{text}}\left( f_{\text{text}}\left( \mathbf{y} \right) \right) \right)$, and $\mathbf{V} = \text{Concat}\left( \mathcal{V}_{\text{image}}\left( f_{\text{image}}\left( \mathbf{z} \right) \right), \mathcal{V}_{\text{text}}\left( f_{\text{text}}\left( \mathbf{y} \right) \right) \right)$, where $\mathcal{Q}_{\text{image}}\left(\cdot\right)$, $\mathcal{K}_{\text{image}}\left(\cdot\right)$, and $\mathcal{V}_{\text{image}}\left(\cdot\right)$ denote the linear projections for image embeddings, and $\mathcal{Q}_{\text{text}}\left(\cdot\right)$, $\mathcal{K}_{\text{text}}\left(\cdot\right)$, and $\mathcal{V}_{\text{text}}\left(\cdot\right)$ denote the linear projections for text embeddings. $\text{Concat}\left( \cdot \right)$ is sequence-wise concatenation. The joint self-attention map $\mathbf{A}$ is computed via: $\mathbf{A} = \left( 
        \begin{array}{cc}
            \mathbf{A}^{\text{II}} & \mathbf{A}^{\text{IT}} \\
            \mathbf{A}^{\text{TI}} & \mathbf{A}^{\text{TT}}
        \end{array}
   \right) 
   = \text{softmax}\left( \frac{\mathbf{Q}\mathbf{K}^T}{\sqrt{d}} \right)$, where $d$ is channel dimension. For simplicity, we omit the denoising timestep $t$. In this work, we focus on $\mathbf{A}^{\text{IT}}$, where we denote the cross-attention map that corresponds to the $i$-th text token as $\mathbf{A}^{\text{IT}}_{\mathbf{y}_i}$.

\section{{Approach}}
\label{sec:approach}

The core innovation of our approach is Cross-Timestep Self-Calibration (\textsc{CTCal}), which leverages reliable text-image alignments learned at small timesteps to calibrate the learning at larger timesteps. This section is organized as follows. Sec.~\ref{sec:overview} provides an overview of the training paradigm. Sec.~\ref{sec:ctcal} details a comprehensive description of the \textsc{CTCal} method. Sec.~\ref{sec:training_strategy} outlines our training strategy.

\subsection{Overview}
\label{sec:overview}

Fig.~\ref{fig:ctcal} illustrates our proposed training paradigm, which deviates from the conventional training approach. Given a real image $\mathbf{I}_{\text{real}}$, a text prompt $\mathbf{y}$, and a Gaussian noise $\mathbf{\epsilon}$, we sample two distinct timesteps, referred to as $t_{\text{stu}}$ and $t_{\text{tea}}$, with $t_{\text{tea}} < t_{\text{stu}}$. Besides predicting the corresponding noises $\mathbf{\epsilon}_{\text{stu}}$ and $\mathbf{\epsilon}_{\text{tea}}$, we also extract and store the cross-attention maps $\mathbf{A}_{\text{stu}}$ and $\mathbf{A}_{\text{tea}}$, computed during the forward process of the denoising network. Notably, both $\mathbf{A}_{\text{stu}}$ and $\mathbf{A}_{\text{tea}}$ are extracted from the same diffusion model, which is currently being fine-tuned. Unlike constructing a separate and fixed pre-trained model for extracting $\mathbf{A}_{\text{tea}}$, our design allows $\mathbf{A}_{\text{tea}}$ to benefit from the learning on newly introduced high-quality data. The aggregated map $\mathbf{A}_{\text{stu / tea}}\in \mathbb{R}^{H\times W\times n}$ consists of $n$ spatial attention maps, each associated with a token of the text prompt. More details on the workflow for processing cross-attention maps are provided in the supplementary material.

Notably, we restrict the optimization to the denoising network associated with timestep $t_{\text{stu}}$, and truncate the gradient of $\mathbf{A}_{\text{tea}}$. Furthermore, we leverage the cross-attention maps $\mathbf{A}_{\text{tea}}$ derived from smaller timestep $t_{\text{tea}}$ as a guide for learning the cross-attention maps $\mathbf{A}_{\text{stu}}$ from larger timestep $t_{\text{stu}}$. This approach explicitly transfers knowledge about text-image correspondence, which is more accurately captured at smaller timesteps, to enhance the learning at larger timesteps. The optimization objective is redefined as:
\begin{equation}
\begin{aligned}
\label{eq:ctcal-base}
\mathcal{L} &= \mathcal{L}_{\text{diffusion}} + \mathcal{L}_{\text{\textsc{CTCal}}} \\
    &= {\mathcal{D}\left( \mathbf{\epsilon}, \epsilon_{\theta}\left( \mathtt{Add\_Noise}\left( \mathbf{I}_{\text{real}}, \mathbf{\epsilon}, t_{\text{stu}} \right), \mathbf{y}, t_{\text{stu}} \right) \right)}\\ &+ {\mathcal{D}\left( \mathbf{A}_{\text{stu}}, \mathbf{A}_{\text{tea}} \right)}.
\end{aligned}
\end{equation}

\begin{figure}
\centering
\setlength{\belowcaptionskip}{-0.3cm}
\setlength{\abovecaptionskip}{0.1cm}
\includegraphics[width=1.\linewidth]{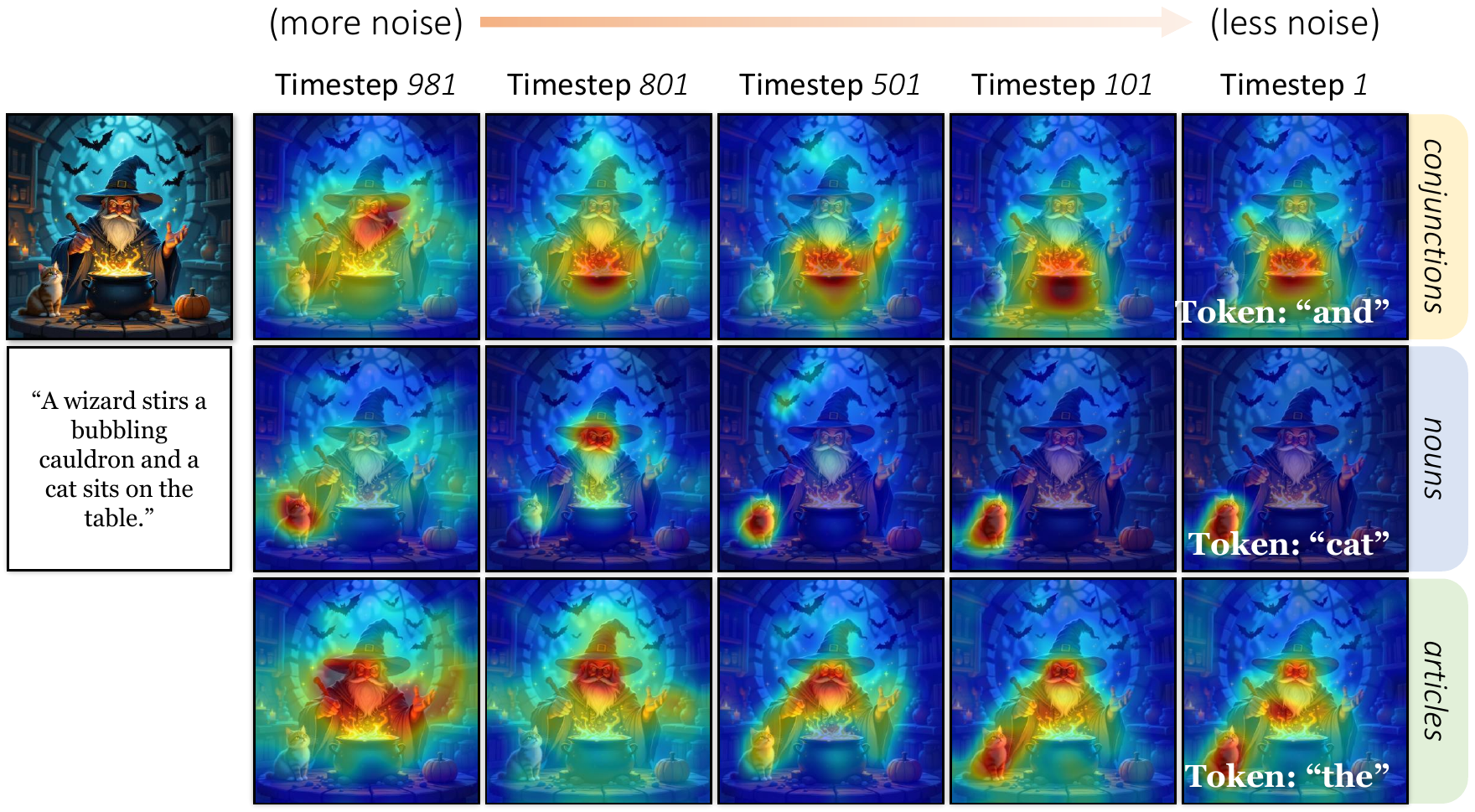}
\caption{\textbf{Investigation on cross-attention maps categorized by part-of-speech.} Cross-ttention maps for noun tokens (\emph{e.g.}, ``cat'') typically encode clear spatial semantic information, while those for articles (\emph{e.g.}, ``the'') and conjunctions (\emph{e.g.}, ``and'') seemingly lack a significant conveyance.}
\label{fig:d_attn_map_selection}
\end{figure}

\subsection{\textsc{CTCal}}
\label{sec:ctcal}

This section provide a detailed explanation of Cross-Timestep Self-Calibration (\textsc{CTCal}), which consists of the following three carefully designed components.

\noindent \textbf{Part-of-speech-based cross-attention map selection strategy.} Given an aggregated cross-attention map $\mathbf{A}_{\text{stu / tea}}\in \mathbb{R}^{H\times W\times n}$, consisting of $n$ spatial attention maps. However, as shown in Fig.~\ref{fig:d_attn_map_selection}, not all tokens yield attention maps that encapsulate meaningful spatial semantic information. For example, tokens representing articles (\emph{e.g.}, ``the'') and conjunctions (\emph{e.g.}, ``and'') may not convey meaningful spatial semantics. Overemphasis on them could potentially degrade the performance.

To rectify this, we propose a part-of-speech-based cross-attention map selection strategy that only extracts and utilizes the attention maps associated with tokens likely to convey significant spatial semantics, specifically, nouns (denoting objects or entities). We reformulate $\mathcal{L}_{\textsc{CTCal}}$ as follows:
\begin{equation}
\begin{aligned}
\label{eq:ctcal-selection}
    \mathcal{L}_{\text{\textsc{CTCal}}} = \frac{1}{N_{\text{noun}}}\sum_{
    \mathbf{y}_i\in\mathcal{Y}_{\text{noun}}
    } \mathcal{D}\left( \mathbf{A}_{\text{stu},\mathbf{y}_i}, \mathbf{A}_{\text{tea},\mathbf{y}_i} \right),
\end{aligned}
\end{equation}
where $\mathcal{Y}_{\text{noun}}$ denotes the set of noun tokens, and $N_{\text{noun}}$ is the number of noun tokens. By restricting the selection of attention maps to this subset, \textsc{CTCal} prioritizes tokens that contribute most directly to spatial understanding.

\noindent \textbf{Pixel-semantic space joint optimization.} To achieve alignment between $\mathbf{A}_{\text{stu}}$ and $\mathbf{A}_{\text{tea}}$, we propose a joint optimization paradigm that simultaneously considers both pixel-level and semantic-level representations. Empirical evidence substantiates the superior performance of this methodology as against an exclusive emphasis on either of the two constituents. We redefine the $\mathcal{L}_{\textsc{CTCal}}$ as follows:
\begin{equation}
\begin{aligned}
\label{eq:ctcal-pixel-semantic}
\mathcal{L}_{\text{\textsc{CTCal}}} &= \frac{1}{N_{\text{noun}}}\sum_{\mathbf{y}_i\in\mathcal{Y}_{\text{noun}}} \lambda_1 {\mathcal{D}\left( \mathbf{A}_{\text{stu},\mathbf{y}_i}, \mathbf{A}_{\text{tea},\mathbf{y}_i} \right)}\\ &+ \lambda_2 {\mathcal{D}\left( f_{\text{attn}}\left( \mathbf{A}_{\text{stu},\mathbf{y}_i} \right), f_{\text{attn}}\left( \mathbf{A}_{\text{tea},\mathbf{y}_i} \right) \right)},
\end{aligned}
\end{equation}
where $f_{\text{attn}}\left(\cdot\right)$ denotes the feature encoder that projects attention maps to their respective semantic representations. A notable concern is the potential overfitting of $f_{\text{attn}}\left(\cdot\right)$ during training, which may instigate mode collapse, causing $f_{\text{attn}}\left(\cdot\right)$ to project all attention maps to identical encodings.

To mitigate this risk, we devise a lightweight autoencoder, composed of an encoder $f_{\text{attn}}^{\text{enc}}(\cdot)$ and a decoder $f_{\text{attn}}^{\text{dec}}(\cdot)$. We apply a reconstruction proxy task as a preventive measure against overfitting:
\begin{equation}
\begin{aligned}
\label{eq:ctcal-pixel-semantic-autoencoder}
\mathcal{L}_{\text{\textsc{CTCal}}} &= \frac{1}{N_{\text{noun}}}\sum_{\mathbf{y}_i\in\mathcal{Y}_{\text{noun}}} \lambda_1 {\mathcal{D}\left( \mathbf{A}_{\text{stu},\mathbf{y}_i}, \mathbf{A}_{\text{tea},\mathbf{y}_i} \right)}\\ &+ \lambda_2 {\mathcal{D}\left( f_{\text{attn}}^{\text{enc}}\left( \mathbf{A}_{\text{stu},\mathbf{y}_i} \right), f_{\text{attn}}^{\text{enc}}\left( \mathbf{A}_{\text{tea},\mathbf{y}_i} \right) \right)}\\
&+ \lambda_3 {
    \mathcal{D}\left( f_{\text{attn}}^{\text{dec}}\left( f_{\text{attn}}^{\text{enc}}\left( \mathbf{A}_{\text{tea},\mathbf{y}_i} \right)\right), \mathbf{A}_{\text{tea},\mathbf{y}_i} \right)
    }.
\end{aligned}
\end{equation}
The detailed architecture of the proposed autoencoder is presented in the supplementary material.

\textbf{Subject response alignment regularization.} \textsc{CTCal} focuses on spatial alignment but may suffer from the imbalanced cross-attention responses among subjects, \emph{i.e.}, subjects with higher responses may overshadow those with lower responses, resulting in the latter being ineffectively rendered in the generated image. Therefore, we introduce subject response alignment regularization, which aligns the cross-attention responses of all subjects to that of the subject with the highest response:
\begin{equation}
\begin{aligned}
\label{eq:subject-response-alignment-regularization}
    \mathcal{R}_{\text{subject}} =\frac{1}{N_{\text{noun}}}\sum_{\mathbf{y}_i\in\mathcal{Y}_{\text{noun}}} \text{ReLU}\left(\mathcal{S}_{\text{attn}} - \max\left( \mathbf{A}_{\text{stu},\mathbf{y}_i} \right) - \tau \right),\\
\end{aligned}
\end{equation}
where $\mathcal{S}_{\text{attn}} = \max_{\mathbf{y}_i\in\mathcal{Y}_{\text{noun}}}\max\left( \mathbf{A}_{\text{stu},\mathbf{y}_i} \right)$. Following \citep{chefer2023attend}, we use $\max\left( \mathbf{A}_{\text{stu},\mathbf{y}_i} \right)$ to characterize the attention response level of subject token $\mathbf{y}_i$. $\tau$ denotes the threshold, and $\text{ReLU}(\cdot)$ ensures that only pairs with differences exceeding $\tau$ contribute to the loss. This design ensures the comparability of responses across different subjects, while effectively preventing the attention responses from increasing unconstrainedly during training. 

\begin{figure*}[ht]
\centering
\setlength{\belowcaptionskip}{-0.3cm}
\setlength{\abovecaptionskip}{0.1cm}
\includegraphics[width=0.913\linewidth]{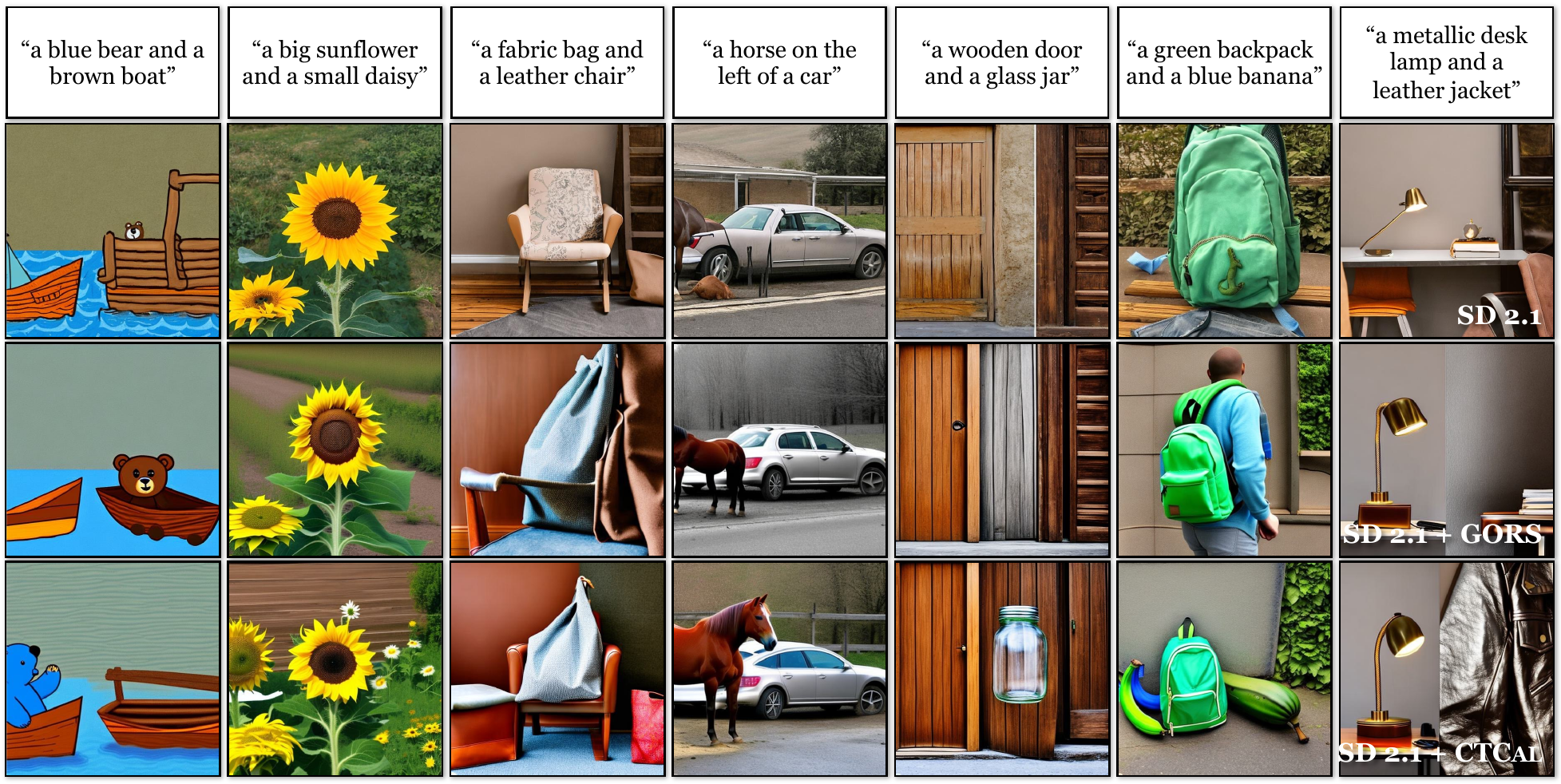}
\includegraphics[width=0.913\linewidth]{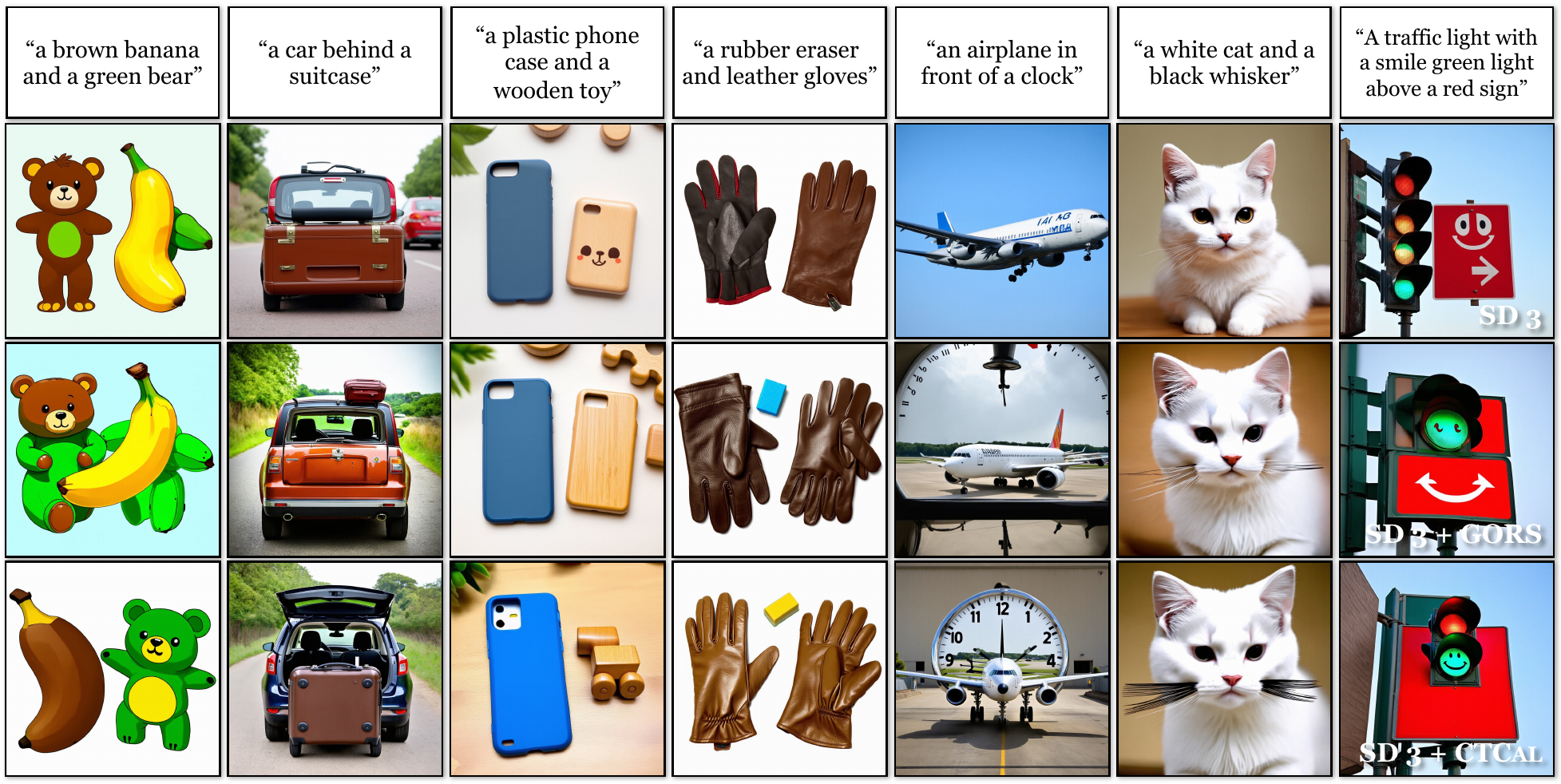}
\caption{\textbf{Qualitative comparison on SD 2.1 and SD 3.} \textsc{CTCal} demonstrates a marked improvement in the fine-grained alignment of generated images with the corresponding text prompts. Each image is generated with the same prompt and random seed for all methods.}
\label{fig:e_exp_sd3}
\end{figure*}

\noindent \textbf{\textsc{CTCal}.} In summary, $\mathcal{L}_{\textsc{CTCal}}$ is ultimately represented as:
\begin{equation}
\begin{aligned}
\label{eq:ctcal}
\mathcal{L}&_{\text{\textsc{CTCal}}} = \frac{1}{N_{\text{noun}}}\sum_{\mathbf{y}_i\in\mathcal{Y}_{\text{noun}}} \lambda_1 \underbrace{\mathcal{D}\left( \mathbf{A}_{\text{stu},\mathbf{y}_i}, \mathbf{A}_{\text{tea},\mathbf{y}_i} \right)}_{\text{Pixel-level loss}}\\ &+ \lambda_2 \underbrace{\mathcal{D}\left( f_{\text{attn}}^{\text{enc}}\left( \mathbf{A}_{\text{stu},\mathbf{y}_i} \right), f_{\text{attn}}^{\text{enc}}\left( \mathbf{A}_{\text{tea},\mathbf{y}_i} \right) \right)}_{\text{Semantic-level loss}} \\
    &+ \lambda_3 \underbrace{
    \mathcal{D}\left( f_{\text{attn}}^{\text{dec}}\left( \mathbf{A}_{\text{tea},\mathbf{y}_i} \right), \mathbf{A}_{\text{tea},\mathbf{y}_i} \right)
    }_{\text{Reconstruction proxy task}} + \lambda_4\underbrace{\mathcal{R}_{\text{subject}}}_{\text{Regularization}},
\end{aligned}
\end{equation}
where $\lambda_1$, $\lambda_2$, $\lambda_3$, and $\lambda_4$ are the tradeoff parameters. Further details on the tradeoff parameters settings are provided in the supplementary material.

\begin{table*}
\setlength{\belowcaptionskip}{-0.2cm}
\setlength{\abovecaptionskip}{0cm}
\begin{center}
\resizebox{0.925\linewidth}{!}{
\begin{threeparttable}
\begin{tabular}{l|cccccccc}
\toprule
\multicolumn{1}{c|}{\bf Methods} & \makecell{\bf Color\\ B-VQA} & \makecell{\bf Shape\\ B-VQA} & \makecell{\bf Texture\\ B-VQA} & \makecell{\bf 2D-Spatial\\ UniDet} & \makecell{\bf 3D-Spatial\\ UniDet} & \makecell{\bf Numeracy\\ UniDet} & \makecell{\bf Non-Spatial\\ Share-CoT} & \makecell{\bf Complex\\ 3-in-1} \\
\midrule
\midrule
SD 1.4 $\text{\citep{rombach2022high}}$ & 0.3765 & 0.3576 & 0.4156 & 0.1246 & 0.3030 & 0.4456 & 0.7487 & 0.3080 \\
SD 2.1 $\text{\citep{rombach2022high}}$ & 0.5065 & 0.4221 & 0.4922 & 0.1342 & 0.3230 & 0.4582 & 0.7567 & 0.3386 \\
\midrule
SD 2.1 + CD $\text{\citep{liu2022compositional}}$ & 0.4063 & 0.3299 & 0.3645 & 0.0800 & 0.2847 & 0.4272 & 0.6927 & 0.2898 \\
SD 2.1 + SD $\text{\citep{feng2023training}}$ & 0.4990 & 0.4218 & 0.4900 & 0.1386 & 0.3224 & 0.4557 & 0.7560 & 0.3355 \\
SD 2.1 + AE $\text{\citep{chefer2023attend}}$ & 0.6400 & 0.4517 & 0.5963 & 0.1455 & 0.3222 & 0.4773 & 0.7593 & 0.3401 \\
\midrule
SD 2.1 + GORS\tnote{1} & 0.6603 & 0.4785 & 0.6287 & 0.1815 & 0.3572 & 0.4830 & 0.7637 & 0.3328 \\
SD 2.1 + GORS\tnote{2} & 0.6426 & 0.4864 & 0.6319 & 0.1775 & 0.3475 & 0.4856 & 0.7621 & 0.3371\\
\midrule
SD 2.1 + \textsc{CTCal} & {\bf 0.7233} & {\bf 0.5149} & {\bf 0.6754} & {\bf 0.2142} & {\bf 0.3862} & {\bf 0.5084} & {\bf 0.7723} & {\bf 0.3403} \\
\bottomrule
\toprule
SD XL $\text{\citep{podell2024sdxl}}$ & 0.5879 & 0.4687 & 0.5299 & 0.2133 & 0.3566 & 0.4991 & 0.7673 & 0.3237 \\
Pixart-$\alpha$-ft $\text{\citep{chen2024pixart}}$ & 0.6690 & 0.4927 & 0.6477 & 0.2064 & 0.3901 & 0.5032 & 0.7747 & 0.3433 \\
DALL-E 3 $\text{\citep{betker2023improving}}$ & 0.7785 & 0.6205 & 0.7036 & 0.2865 & 0.3744 & 0.5926 & 0.7853 & 0.3773 \\
FLUX-schnell $\text{\citep{flux2024}}$ & 0.7407 & 0.5718 & 0.6922 & 0.2863 & 0.3866 & 0.6185 & 0.7809 &  0.3703 \\
SD 3 (2B) $\text{\citep{esser2024scaling}}$ & 0.8132 & 0.5885 & 0.7334 & 0.3200 & 0.4084 & 0.6174 & 0.7782 &  0.3771 \\
\midrule
SD 3 (2B) + CORS\tnote{2} & 0.8236 & 0.5833 & 0.7398 & 0.3232 & 0.4033 & 0.6280 & 0.7708 & 0.3739\\
\midrule
SD 3 (2B) + \textsc{CTCal} & {\bf 0.8443} & {\bf 0.5968} & {\bf 0.7581} & {\bf 0.3476} & {\bf 0.4117} & {\bf 0.6292} & {\bf 0.7867} & {\bf 0.3814} \\
\bottomrule
\end{tabular}
\begin{tablenotes}   
\footnotesize         
\item[1] The results are sourced from the original paper \citep{huang2025t2i}.
\item[2] The results are derived from our reimplementation using the text-image dataset we constructed. 
\end{tablenotes} 
\end{threeparttable}}
\end{center}
\caption{\textbf{Objective evaluation on T2I-CompBench++.} Top: conventional text-to-image diffusion models, inference-time optimization approaches, supervised fine-tuning methods, and our method. Bottom: advanced text-to-image diffusion models, supervised fine-tuning methods, and our method. \textsc{CTCal} exhibits the superior performance in attribute binding, object relationships, counting, and complex compositions, highlighting the advanced capability for compositional generation.}
\label{tab:t2i-compbench-exp}
\end{table*}

\subsection{Training strategy}
\label{sec:training_strategy}

\noindent \textbf{Training timestep sampling strategy.} For $t_{\text{stu}}$, we strictly adhere to the inherent timestep sampling protocol that has been established by the text-to-image diffusion models during the training stage. For $t_{\text{tea}}$, we empirically set $t_{\text{tea}} = 0$ for classical text-to-image diffusion models (\emph{e.g.}, Stable Diffusion 2.1). This particular selection corresponds to the regime with minimal noise.

It is critical to highlight that, in contrast to classical methods that rely on uniform timestep sampling, contemporary cutting-edge models (\emph{e.g.}, SD 3) have integrated non-uniform timestep samplers, such as the logit-normal sampler. Consequently, a reevaluation of timestep priority based on the sampling distribution becomes crucial for identifying $t_{\text{tea}}$. Naively setting $t_{\text{tea}}=0$ may thus degrade performance. Further discussion and implementation details specific to SD 3 are provided in the supplementary material.

\noindent \textbf{Timestep-aware adaptive weighting.} To augment the potency of $\mathcal{L}_{\textsc{CTCal}}$, we introduce a timestep-aware adaptive weighting scheme. Specifically, during the initial stages (\emph{i.e.}, with less noise) of the diffusion process, the diffusion loss predominantly governs the alignment between textual and visual modalities, rendering a lower contribution from $\mathcal{L}_{\textsc{CTCal}}$. In contrast, at larger timesteps (\emph{i.e.}, with more noise), the model relies more heavily on $\mathcal{L}_{\textsc{CTCal}}$. 

We formalize this intuition using a simple yet effective linear weighting function that scales the influence of $\mathcal{L}_{\textsc{CTCal}}$ according to the current diffusion timestep:
\begin{equation}
\begin{aligned}
\label{eq:ctcal-weight}
    \mathcal{L} = \mathcal{L}_{\text{diffusion}} + \lambda_t \mathcal{L}_{\text{\textsc{CTCal}}}, \quad \text{where}\ \lambda_t = \frac{t_{\text{stu}}}{T_{\text{train}}},
\end{aligned}
\end{equation}
where $t_{\text{stu}}$ is the current timestep and $T_{\text{train}}$ is the total number of diffusion steps during training. Thus, $\lambda_t$ increases linearly with $t_{\text{stu}}$, assigning greater emphasis to $\mathcal{L}_{\textsc{CTCal}}$ as the process advances. This adaptive scheme enables the model to balance both objectives throughout training, facilitating stable convergence and improved performance.

\begin{table*}
\setlength{\belowcaptionskip}{-0.2cm}
\setlength{\abovecaptionskip}{0.2cm}
    \centering 
    \begin{minipage}[t]{0.7\textwidth} 
        \centering
        \resizebox{0.915\textwidth}{!}{\begin{tabular}{l|ccccccc}
    \toprule
    \multicolumn{1}{c|}{\bf Methods} & \makecell{\bf Overall} & \makecell{\bf Single\\ \bf object} & \makecell{\bf Two\\ \bf object} & \makecell{\bf Counting} & \makecell{\bf Colors} & \makecell{\bf Position} & \makecell{\bf Color\\ \bf attribution} \\
    \midrule
    \midrule
    SD 2.1 $\text{\citep{rombach2022high}}$ & 0.50 & 0.98 & 0.51 & 0.44 & 0.85 & 0.07 & 0.17 \\
    \midrule
    SD 3 (2B) \citep{esser2024scaling} & 0.62 & 0.98 & 0.74 & 0.63 & 0.67 & 0.34 & 0.36 \\
    SD 3 (2B) + \textsc{CTCal} & {\bf 0.69} & {\bf 0.99} & {\bf 0.85} & {\bf 0.70} & {\bf 0.79} & {\bf 0.38} & {\bf 0.42} \\
    \bottomrule
    \end{tabular}}
    \caption{\textbf{Objective evaluation on GenEval.} \textsc{CTCal} improves performance across all categories.}
    \label{tab:geneval-exp}
    \end{minipage}
    \hfill
    \begin{minipage}[t]{0.29\textwidth}
        \centering
        \resizebox{1\textwidth}{!}{\begin{tabular}{l|c|c}
        \toprule
        \multicolumn{1}{c|}{\bf Methods} & \multicolumn{1}{c}{\makecell{\bf SD 2.1\\ User study}} & \multicolumn{1}{c}{\makecell{\bf SD 3\\ User study}} \\
        \midrule
        \midrule
        SD 2.1 / SD 3 & 4.17\% & 24.17\% \\
        + GORS & 19.17\% & 21.67\% \\
        + \textsc{CTCal} & {\bf 76.67\%} & {\bf 54.17\%} \\
        \bottomrule
        \end{tabular}}
        \caption{\textbf{User study.}}
\label{tab:user-study}
    \end{minipage}
\end{table*}

\section{Experiments}
\label{sec:experiments}

\subsection{Experimental settings}
\label{sec:experimental_settings}

\noindent \textbf{Implementation details.} \textsc{CTCal} is a model-agnostic training paradigm that can be seamlessly incorporated into prevailing text-to-image diffusion frameworks. To validate the efficacy and generalizability of \textsc{CTCal}, we integrate it with two highly recognized diffusion models: Stable Diffusion 2.1 (SD 2.1) \citep{rombach2022high} and Stable Diffusion 3 (SD 3) \citep{esser2024scaling}. \textsc{CTCal} is implemented within the Diffusers codebase, employing Low-Rank Adaptation (LoRA) to fine-tune both the self-attention layers of the text encoder and the attention layers of the denoising network. We use \emph{Stanza} for part-of-speech analysis to extract nouns from the given text prompts. $\mathcal{D}(\cdot)$ is implemented as the mean squared error loss function for \textsc{CTCal}. We conduct a comprehensive evaluation of \textsc{CTCal} utilizing two widely recognized benchmarks: T2I-CompBench++ \citep{huang2025t2i} and GenEval \citep{ghosh2023geneval}. More parameter setting, training and evaluation details are provided in the supplementary material. 

\noindent \textbf{Datasets.} Current mainstream text-to-image generation models \citep{balaji2022ediffi,xue2023raphael,betker2023improving,gu2023matryoshka,podell2024sdxl,li2024hunyuan,chen2024pixart,flux2024,esser2024scaling,xie2025sana} are predominantly trained on proprietary datasets, resulting in a paucity of open-source, high-quality text-image pair datasets within the research community. To address this limitation, we adopt the dataset construction method proposed by \citep{huang2025t2i}, which utilizes a reward-driven sample selection strategy to curate training dataset. Specifically, we utilize the text prompt dataset from \citep{huang2025t2i}, which comprises 700 prompts per category. For each prompt, we generate $k$ images using the target text-to-image diffusion model, thereby forming a set of candidate text-image pairs. Each candidate pair is then evaluated using the scoring metric introduced in \citep{huang2025t2i}. We subsequently select the top-$n$ pairs with the highest scores from each candidate set to fine-tune the diffusion model. In our experiments, for each category in \citep{huang2025t2i}, we set $k=100$, $n=10,000$ for SD 2.1, and $k=30$, $n=10,000$ for SD 3.

\subsection{Qualitative comparison}
\label{sec:qualitative_comparison}

Fig.~\ref{fig:e_exp_sd3} present a comparative analysis of our method against supervised fine-tuning approach (\emph{i.e.}, GORS) using identical text prompts and random seeds on Stable Diffusion 2.1 (SD 2.1) and Stable Diffusion 3 (SD 3). GORS leverages synthesized text-image data meticulously selected based on reward functions as detailed in Sec.~\ref{sec:experimental_settings}, and adopts the standard diffusion loss for model fine-tuning. Building on this, our approach incorporates \textsc{CTCal}. 

As illustrated in Fig~\ref{fig:e_exp_sd3}, SD 2.1 struggles with compositional text-to-image synthesis. While GORS demonstrates improvements, it still exhibits limitations in accurately rendering uncommon concepts. For instance, GORS enhances the depiction of ``blue'' but fails to render a ``blue banana''. In contrast, our method successfully synthesizes such challenging compositions. Fig.~\ref{fig:e_exp_sd3} shows that SD 3, benefiting from extensive high-quality datasets and advanced architectures, already achieve strong performance on text-guided image generation. Nonetheless, \textsc{CTCal} further enhances performance beyond this baseline.

\subsection{Quantitative comparison}
\label{sec:quantitative_comparison}

\noindent \textbf{Objective evaluation.} Table~\ref{tab:t2i-compbench-exp} presents quantitative results on T2I-CompBench++ \citep{huang2025t2i}. \textsc{CTCal} demonstrates substantial improvements over existing text-to-image diffusion models in attribute binding, object relationships, counting, and complex compositions, including diffusion-based (SD 2.1) and flow-based method (SD 3). Furthermore, \textsc{CTCal} outperforms inference-time optimization and supervised fine-tuning methods, confirming the effectiveness and generalizability.

Notably, noun-token-based \textsc{CTCal} still enhances performance on the dimensions of action and positional relationship. This is primarily attributed to the supervision of accurate subject rendering, which also improves the ability to understand and learn from training images. Furthermore, rendering the subject at the correct position partially integrates positional and action information. The manifestation of both positional and action information depends on the subject, which serves as the foundation.

To mitigate potential biases introduced by employing evaluation metrics as rewards during dataset construction, we further report cross-benchmark validation results in Table~\ref{tab:geneval-exp}. Unlike the protocol in \citep{huang2025t2i}, which fine-tunes LoRA parameters for specific categories, we aggregate text-image pairs of all categories (80,000 pairs) for joint fine-tuning. As shown in Table~\ref{tab:geneval-exp}, evaluation on GenEval \citep{ghosh2023geneval} indicates that \textsc{CTCal} consistently improves performance across all categories, further substantiating the robustness.

\noindent \textbf{User study.} A subjective user study is conducted with 12 volunteers, 6 of whom have expertise in image processing. Participants are asked to select the most visually appealing and semantically faithful images, with 10 questions per participant. We record the voting results and present the statistics in Table~\ref{tab:user-study}. Our method performs favorably against the other methods.

\subsection{Ablation study}
\label{sec:ablation_study}

We perform the ablation study on the {Color} and {2D-Spatial} categories of T2I-CompBench++ to systematically evaluate the effectiveness of our design. We define as follows: (\textsc{a}) denotes the naive constraint of $\mathbf{A}_{\text{tea}}$ and $\mathbf{A}_{\text{stu}}$, (\textsc{b}) introduces a part-of-speech-based cross-attention map selection strategy based on (\textsc{a}), (\textsc{c}) introduces pixel-semantic space joint optimization based on (\textsc{b}), (\textsc{d}) introduces subject response alignment regularization based on (\textsc{c}), and (\textsc{e}) introduces timestep-aware adaptive weighting based on (\textsc{d}). (\textsc{e}) is the final version of \textsc{CTCal}.

As shown in Table~\ref{tab:ablation-study}, (\textsc{a}) even decreases performance due to considering the attention maps that do not contain spatial semantic information. With the priority given to noun tokens by the part-of-speech-based cross-attention map selection strategy, (\textsc{b}) demonstrates its effectiveness, significantly enhancing performance. The Pixel-semantic space joint optimization, Subject response alignment regularization, and Timestep-aware adaptive weighting further optimize the performance of \textsc{CTCal}.

\begin{table}
\small
\centering
\setlength{\belowcaptionskip}{-0.2cm}
\setlength{\abovecaptionskip}{0.2cm}
\resizebox{0.98\linewidth}{!}{
\begin{tabular}{l|ll}
\toprule
\multicolumn{1}{c|}{\bf Methods} & \multicolumn{1}{c}{\makecell{\bf Color\\ B-VQA}} & \multicolumn{1}{c}{\makecell{\bf 2D-Spatial\\ UniDet}} \\
\midrule
\midrule
SD 2.1 & \multicolumn{1}{c}{0.5065} & \multicolumn{1}{c}{0.1342} \\
+ GORS (baseline) & \multicolumn{1}{c}{0.6426} & \multicolumn{1}{c}{0.1775} \\
\midrule
+ \textsc{CTCal} (\textsc{a}) & 0.6286 (\textcolor{gray}{-2.18\%}) & 0.1693 (\textcolor{gray}{-4.62\%}) \\
+ \textsc{CTCal} (\textsc{b}) & 0.6897 (\textcolor{red}{+7.33\%}) & 0.1972 (\textcolor{red}{+11.10\%}) \\
+ \textsc{CTCal} (\textsc{c}) & 0.6992 (\textcolor{red}{+8.81\%}) & 0.2021 (\textcolor{red}{+13.86\%}) \\
+ \textsc{CTCal} (\textsc{d}) & 0.7148 (\textcolor{red}{+11.24\%}) & 0.2095 (\textcolor{red}{+18.03\%}) \\
\midrule
+ \textsc{CTCal} (\textsc{e}) & {\bf 0.7233} ({\bf \textcolor{red}{+12.56\%}}) & {\bf 0.2142} ({\bf \textcolor{red}{+20.68\%}})  \\
\bottomrule
\end{tabular}}
\caption{{\bf Ablation study} on Stable Diffusion 2.1.}
\label{tab:ablation-study}
\end{table}

\begin{table*}
\setlength{\belowcaptionskip}{-0.1cm}
\setlength{\abovecaptionskip}{0.2cm}
    \centering
    \begin{minipage}[t]{0.695\textwidth}
        \vspace{0pt}
        \centering
        \includegraphics[width=1.\textwidth]{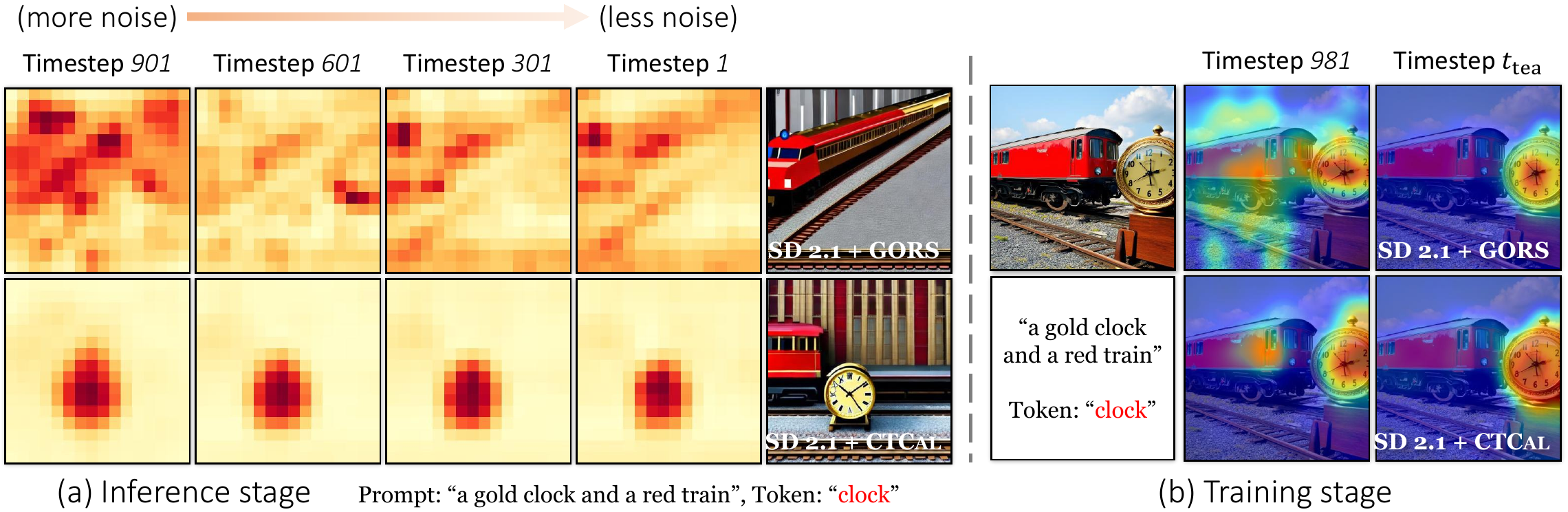}
        \captionof{figure}{\textbf{Visualization of cross-attention maps} extracted from the fine-tuned models.}
        \label{fig:e_attn_cal}
    \end{minipage}
    \hfill
    \begin{minipage}[t]{0.285\textwidth}
        \vspace{0pt}
        \centering
        \resizebox{0.95\linewidth}{!}{
        \begin{tabular}{l|cc}
        \toprule
        \multicolumn{1}{c|}{\bf Methods} & \makecell{\bf Color\\ B-VQA} \\
        \midrule
        \midrule
        SD 2.1 & \multicolumn{1}{c}{0.5065} \\
        + \textsc{CTCal} ($0 \le t_{\text{tea}} < t_{\text{stu}}$) & {0.7028}\\
        + \textsc{CTCal} (Ours) & {\bf 0.7233}  \\
        \bottomrule
        \toprule
        \multicolumn{1}{c|}{\bf Methods} & \makecell{\bf 2D-Spatial\\ UniDet} \\
        \midrule
        \midrule
        SD 2.1 & \multicolumn{1}{c}{0.1342} \\
        + \textsc{CTCal} ($0 \le t_{\text{tea}} < t_{\text{stu}}$)  & {0.2029} \\
        + \textsc{CTCal} (Ours) & {\bf 0.2142} \\
        \bottomrule
        \end{tabular}}
        \caption{\textbf{Objective evaluation on $t_{\text{tea}}$.}}
        \label{tab:t_tea}
    \end{minipage}
\end{table*}

\subsection{More results}

\noindent \textbf{More results on $t_{\text{tea}}$.} As mentioned in Sec.~\ref{sec:ctcal}, \textsc{CTCal} requires $t_{\text{tea}} < t_{\text{stu}}$. For conventional diffusion models (\emph{e.g.}, SD 2.1), we empirically set $t_{\text{tea}}$ to a fixed value of 0. This specific choice corresponds to the scenario with the least noise, representing the most favorable timestep for learning text-image correspondences. To further investigate this, we conducted additional experiments. Instead of setting a fixed $t_{\text{tea}}=0$, we randomly sample $t_{\text{tea}}$ for a given $t_{\text{stu}}$ such that $0 \le t_{\text{tea}} < t_{\text{stu}}$. The quantitative evaluation presented in Table~\ref{tab:t_tea} shows that while the random sampling approach can still enhance the performance, setting $t_{\text{tea}}=0$ remains the superior choice.

\noindent \textbf{More results on part-of-speech-based cross-attention map selection strategy.} As discussed in Sec.~\ref{sec:ctcal}, cross-attention maps corresponding to noun tokens generally encapsulate clear spatial-semantic information. Therefore, \textsc{CTCal} prioritizes the attention maps corresponding to noun tokens. Notably, adjectives, especially those modifying nouns, also demonstrate accurate spatial correspondence. Accordingly, we further discuss tokens with adjectival properties.
Table~\ref{tab:adjective-exp} presents additional quantitative results on the {Color} and {Texture} categories of the T2I-CompBench++ benchmark. These experiments focus on adjective-noun pairs to more rigorously assess attribute rendering accuracy. The results indicate that incorporating alignment on adjective tokens leads to measurable improvements in performance.

\begin{table}
\small
\centering
\setlength{\belowcaptionskip}{-0.3cm}
\setlength{\abovecaptionskip}{0.2cm}
\resizebox{0.78\linewidth}{!}{
\begin{tabular}{l|cc}
\toprule
\multicolumn{1}{c|}{\bf Methods} & \makecell{\bf Color\\ B-VQA} & \makecell{\bf Texture\\ B-VQA} \\
\midrule
\midrule
SD 2.1 & \multicolumn{1}{c}{0.5065} & \multicolumn{1}{c}{0.4922} \\
+ GORS & \multicolumn{1}{c}{0.6426} & \multicolumn{1}{c}{ 0.6319} \\
+ \textsc{CTCal} & {0.7233}  & {0.6754} \\
+ \textsc{CTCal} + adj. token & {\bf 0.7328}  & {\bf 0.6877} \\
\bottomrule
\end{tabular}}
\caption{\textbf{Objective evaluation on \emph{adj.}.}}
\label{tab:adjective-exp}
\end{table}

\noindent \textbf{More results on diversity evaluation.} To assess the diversity of images generated by the proposed method, we conduct a diversity evaluation experiment using the widely adopted Mean LPIPS Distance, where a higher value indicates greater diversity. As shown in Table~\ref{tab:x_diversity_quality}, \textsc{CTCal} improves text-image alignment performance without compromising the diversity of the generated samples.

\noindent \textbf{More results on image quality evaluation.} We present additional quantitative experiments on aesthetic score, which is a reward function for measuring image quality that is independent of text alignment. As shown in the Table~\ref{tab:x_diversity_quality}, our method does not compromise the quality of generated images while improving text-image consistency; on the contrary, it exhibits a moderate improvement in quality. This suggests that the advantage of \textsc{CTCal} in text-image alignment is shared with image quality. Improved text-image alignment can correct semantic confusion and conflicts in the spatial dimension, which enhances the ability to render the correct object in the accurate location, which in turn leads to improved image quality.

\begin{table}
\setlength{\belowcaptionskip}{-0.6cm}
\setlength{\abovecaptionskip}{0.2cm}
\small
\begin{center}
\resizebox{1.\linewidth}{!}{
\begin{tabular}{l|cc|cc}
\toprule
\multicolumn{1}{c|}{\bf Methods} & \makecell{\bf Color\\ M-LPIPS} & \makecell{\bf 2D-Spatial\\ M-LPIPS} & \makecell{\bf Color\\ Aesthetic} & \makecell{\bf 2D-Spatial \\ Aesthetic} \\
\midrule
\midrule
SD 2.1 & \multicolumn{1}{c}{0.637} & \multicolumn{1}{c|}{0.618} & \multicolumn{1}{c}{5.128} & \multicolumn{1}{c}{5.263}  \\
+ GORS & \multicolumn{1}{c}{0.621} & \multicolumn{1}{c|}{0.626} & \multicolumn{1}{c}{5.194} & \multicolumn{1}{c}{5.281} \\
+ \textsc{CTCal} & {0.634}  & {0.623}  & {5.288}  & {5.344} \\
\bottomrule
\end{tabular}}
\caption{\textbf{More results on diversity and quality evaluation.}}
\label{tab:x_diversity_quality}
\end{center}
\end{table}

\noindent \textbf{Visualizations of cross-attention maps.} Fig.~\ref{fig:e_attn_cal} presents visualizations of cross-attention maps generated by the fine-tuned models, depicted separately for (a) inference and (b) training modes. During inference, our method demonstrates accurate and reasonable attention allocation, leading to semantically consistent outputs. Furthermore, compared to GORS, the cross-attention maps derived from \textsc{CTCal} at later timesteps exhibit greater consistency with those at smaller timesteps. This observation shows the efficacy of \textsc{CTCal} and provides empirical support for our insights.

\section{Conclusion}
\label{sec:conclusion}

This study addresses the persistent challenge of precise text-image alignment in text-to-image diffusion models by introducing Cross-Timestep Self-calibration (\textsc{CTCal}). Through a rigorous analysis, we demonstrate that alignment difficulties intensify with increasing diffusion timesteps, underscoring the limitations of conventional diffusion loss. \textsc{CTCal} mitigates this issue by explicitly calibrating the learning at larger timesteps with more noise using the robust text-image alignment established at smaller timesteps with less noise, supplemented by a timestep-aware adaptive weighting mechanism for seamless integration with standard diffusion losses. \textsc{CTCal} is model-agnostic and readily adaptable to a wide array of diffusion-based and flow-based architectures. Extensive evaluation on established benchmarks substantiates the efficacy and generalizability of \textsc{CTCal}, marking a significant advancement toward more accurate and reliable text-to-image generation.

\section*{Acknowledgment} This work is partly supported by the National Key Research and Development Plan (2024YFB3309300), National Natural Science Foundation of China (82441024), the Beijing Natural Science Foundation (L251073), the Research Program of State Key Laboratory of Complex and Critical Software Environment, and the Fundamental Research Funds for the Central Universities.


{
    \small
    \bibliographystyle{ieeenat_fullname}
    \bibliography{main}
}

\end{document}